\documentclass[10pt, a4paper]{article}

\usepackage[]{lrec-coling2024} 
\usepackage{inconsolata}
\usepackage{latexsym}
\usepackage{float}
\usepackage{mathrsfs}
\usepackage{amsthm,amsmath,amssymb}
\usepackage{multirow}
\usepackage[ruled]{algorithm2e}
\usepackage{threeparttable}
\usepackage{array}
\usepackage{booktabs}
\usepackage{bm}
\usepackage{graphicx}
\usepackage{amssymb}
\usepackage{wasysym}
\usepackage{bbm}
\title{Layer-wise Regularized Dropout for Neural Language Models}

\name{
	\begin{tabular}{c}
		Shiwen Ni$^{1}$, 
		Min Yang$^{1\dagger}$\thanks{$^{\dagger}$Corresponding author}, Ruifeng Xu$^2$, Chengming Li$^3$, Xiping Hu$^3$
	\end{tabular}
} 

\address{$^1$Shenzhen Institute of Advanced Technology, Chinese Academy of Sciences \\
	$^2$Harbin Institute of Technology, Shenzhen \\
	$^3$Shenzhen MSU-BIT University \\
	\{sw.ni, min.yang\}@siat.ac.cn;	xuruifeng@hit.edu.cn; \{licm, huxp\}@smbu.edu.cn}

\abstract{
Among the various pre-trained neural language models that are popular today, dropout is already an indispensable regularization technique. To solve the inconsistency between training and inference caused by the randomness of dropout, some studies use consistency training to regularize dropout at the output layer. In this paper, we propose a novel \textbf{L}ayer-wise \textbf{R}egularized \textbf{Drop}out (\textbf{LR-Drop}), which is specially designed for Transformer-based Language models. Specifically, LR-Drop layer-wise regularizes each Transformer layer using the consistency training strategy.  Each training sample passes through the two siamese sub-models sampled by dropout, and then LR-Drop forces the hidden states, multi-head attention matrices, and output distribution of the two siamese sub-models to be consistent. The proposed LR-Drop can be regarded as a “self-distillation” framework, in which each sub-model generated by dropout is the other's “teacher” model and “student” model. Through extensive experiments on 8 natural language understanding datasets, 6 neural machine translation datasets, and 1 abstractive summarization dataset (a total of 15 datasets), we show that LR-Drop achieves superior performances, including state-of-the-art results. 
 \\ \newline  }

\begin{document}

\maketitleabstract

\section{Introduction}
In recent years, pre-trained language models (PLMs) based on the Transformer architecture have revolutionized the field of natural language processing (NLP) by achieving state-of-the-art performance on a wide range of NLP tasks. These models, such as BERT (Bidirectional Encoder Representations from Transformers) \citep{kenton2019bert}, ALBERT (A Lite BERT) \citep{lan2019albert}, XLNet \citep{yang2019xlnet}, RoBERTa \citep{liu2019roberta}, and ELECTRA (Efficiently Learning an Encoder that Classifies Token Replacements Accurately) \citep{clark2019electra}, have demonstrated their effectiveness in tasks such as text classification, named entity recognition, sentiment analysis, machine translation, question answering, and more.

One of the key reasons for the success of these Transformer-based PLMs is their ability to capture contextualized representations of words and sentences. By leveraging the self-attention mechanism, these models can efficiently encode the relationships between different words in a sentence, allowing them to capture long-range dependencies and context. The pre-training stage involves training the models on large amounts of unlabeled text, followed by fine-tuning on specific downstream tasks using labeled data. This transfer learning approach has proven to be highly effective, as the pre-trained models can leverage the knowledge learned from the vast amount of unlabeled data to perform well on a variety of NLP tasks.

To prevent overfitting and improve the generalization ability of PLMs, dropout regularization techniques \citep{srivastava2014dropout} are commonly employed during both the pre-training and fine-tuning stages. Dropout randomly deactivates a portion of the neural units during training, effectively creating an ensemble of sub-models. This ensemble approach helps in reducing over-reliance on specific units and encourages the model to learn more robust and generalizable representations. However, the use of dropout introduces a challenge in terms of inconsistency between training and inference. During training, dropout is applied to create the ensemble, but during inference, the full model without dropout is used, leading to a mismatch in behavior.

Several studies \citep{ma2016dropout, zolna2018fraternal} have highlighted this inconsistency and its potential impact on model performance. They have proposed methods to address this issue by introducing L2 regularization to the hidden unit state. However, the effectiveness of this approach is limited, and it does not fully resolve the inconsistency problem. To tackle this challenge more effectively, recent research \cite{wu2021r} has introduced a novel consistency training method called R-Drop.
R-Drop aims to align the output distributions of identical data samples processed by different sub-models created through dropout. It involves performing two forward passes for each data sample, with each pass handled by a distinct sub-model that randomly deactivates some hidden units. By minimizing the bidirectional Kullback-Leibler (KL) divergence between the output distributions of these two sub-models, R-Drop encourages consistency in the predictions made by the ensemble. This approach provides a more robust and consistent regularization of dropout, addressing the inconsistency issue between training and inference.

In addition to regulating dropout at the output layer, it is also important to ensure consistency in other representations within the PLM. For instance, the multi-head attention mechanism, which is a crucial component of Transformer-based models, typically employs dropout. Previous studies \citep{clark2019does} have shown that the attention weight matrix captures substantial linguistic knowledge. Therefore, it is essential to maintain consistency between the multi-head attention matrices of different sub-models to preserve the learned linguistic knowledge. By extending the principles of R-Drop, we proposed LR-Drop to introduce regular constraints into each Transformer layer of the model.

In particular, we formulate three loss functions to regulate different representations from PLMs layers: 1) the hidden states and 2) multi-head attention matrices extracted from the Transformer layer; 3) the output distributions generated by the prediction layer. The multi-head attention in PLMs typically employs dropout, and previous studies \citep{clark2019does} have demonstrated that the attention weight matrix can acquire a substantial amount of linguistic knowledge, hence we ensure the consistency between the two multi-head attentions. 

To summarize, the main contributions of this paper are as follows:
\begin{itemize}
	\item In this work, we propose the layer-wise regularized dropout (LR-Drop), a simple but effective regularization technique built upon dropout, designed for Transformer-based pre-trained language models. 
	\item For the special structure of Transformer-based pre-trained language models, we are the first to propose Transformer-layer regularization, which includes regularization for hidden states and multi-headed attention.
	\item Our LR-Drop does not introduce additional model parameters and does not change the original architecture of the language model. 
	\item By conducting rigorous experiments on 8 natural language understanding datasets, 6 neural machine translation datasets, and 1 abstractive summarization dataset, we provide evidence that LR-Drop excels in performance, even achieving state-of-the-art results.
\end{itemize}

\section{Related Work}

\subsection{Regularization Methods}
The susceptibility of large and deep neural network models to overfitting is a well-established fact. It has been observed that the most effective models are typically large ones, but they are also paired with appropriate regularization techniques. A plethora of regularization techniques have been suggested to enhance the generalization capacity of these models.
\cite{krogh1992simple} introduced simple weight decay as a regularization technique to improve generalizability . \cite{kang2016shakeout} proposed the Shakeout method, which randomly enhances or inverts the contribution of each cell to the next layer, effectively applying L1 and L2 regularization to the weights. Normalization techniques have also been utilized for regularization by researchers such as \citep{ba2016layer,salimans2016weight,wu2018group}. \cite{hochreiter1995simplifying,poole2014analyzing} found that adding noise can have a regularization effect. Label smoothing, a simple regularization technique particularly effective in the presence of noisy labels, has been explored by \citep{muller2019does,zhang2021delving,li2020regularization}.
Adversarial training, as proposed by \citep{goodfellow2014explaining,zhu2020freelb,ni2021dropattack,ni2022hat4rd,ni2022r} has shown significant improvement in model performance, but it comes at the cost of increased computational effort. Dropout and its derivatives, including Adaptive Dropout by \citep{wan2013regularization,ba2013adaptive,srivastava2014dropout,ni2023masked} have gained popularity due to their effectiveness and compatibility with other regularization techniques. Dropout enables the generation of sub-models with exponentially shared parameters during training, providing powerful regularization capabilities.
\subsection{Knowledge Distillation}
The concept of minimizing the output or parameter distribution between two models is commonly referred to as knowledge distillation \citep{hinton2015distilling,furlanello2018born}. In knowledge distillation, a teacher model and a student model are typically employed, where the student model learns from both the ground truth labels and the teacher model during training. The teacher model serves as a guide for the student model, allowing it to learn the parameters and output distribution of the teacher model. This process can be viewed as the student model distilling knowledge from the teacher model.
In the case of R-Drop \citep{wu2021r}, the generated sub-models can be seen as reciprocal teacher and student models, similar to the concept of self-distillation \citep{mobahi2020self, zhang2019your,zhang2020self}. However, R-Drop only applies self-distillation to the output of the model, without considering the internal representations. On the other hand, our proposed LR-Drop incorporates a \textit{layer-wise self-distillation} approach, similar to the knowledge distillation technique employed in TinyBERT \citep{jiao2020tinybert}. This allows for a more comprehensive knowledge interaction between the sub-models within our LR-Drop framework.
It is important to note that while knowledge distillation is typically used to compress models, the primary objective of LR-Drop is to facilitate mutual learning among the sub-models within the larger model, thereby enhancing overall model performance.

\section{Method: LR-Drop}
This section presents a novel regularization method called LR-Drop, specifically designed for Transformer-based language models. The LR-Drop technique is illustrated in Figure 1, where it is applied to the Transformer-based model. The process begins by inputting a sample $x$ into the model with dropout applied twice, resulting in two output distributions denoted as $P_1$ and $P_2$. Subsequently, the \textit{cross-entropy} loss is calculated using $P_1$, $P_2$, and the hard label $y$:
\begin{equation}\label{key}
	\mathcal{L}_{CE}=-\mathrm{log}_{1}^{w}(y|x)-\mathrm{log}_{2}^{w}(y|x).
\end{equation}
In addition to the losses obtained from the computation with labels, LR-Drop contains three regularization losses, which are Transformer-layer regularization (containing (1) hidden states regularization loss and (2) multi-head attention regularization loss) and (3) output regularization loss. Next we will describe each of these three regularization processes and details.
\subsection{Transformer-layer Regularization}
\begin{figure*}[t]
	\centering
	\includegraphics[width=1\linewidth]{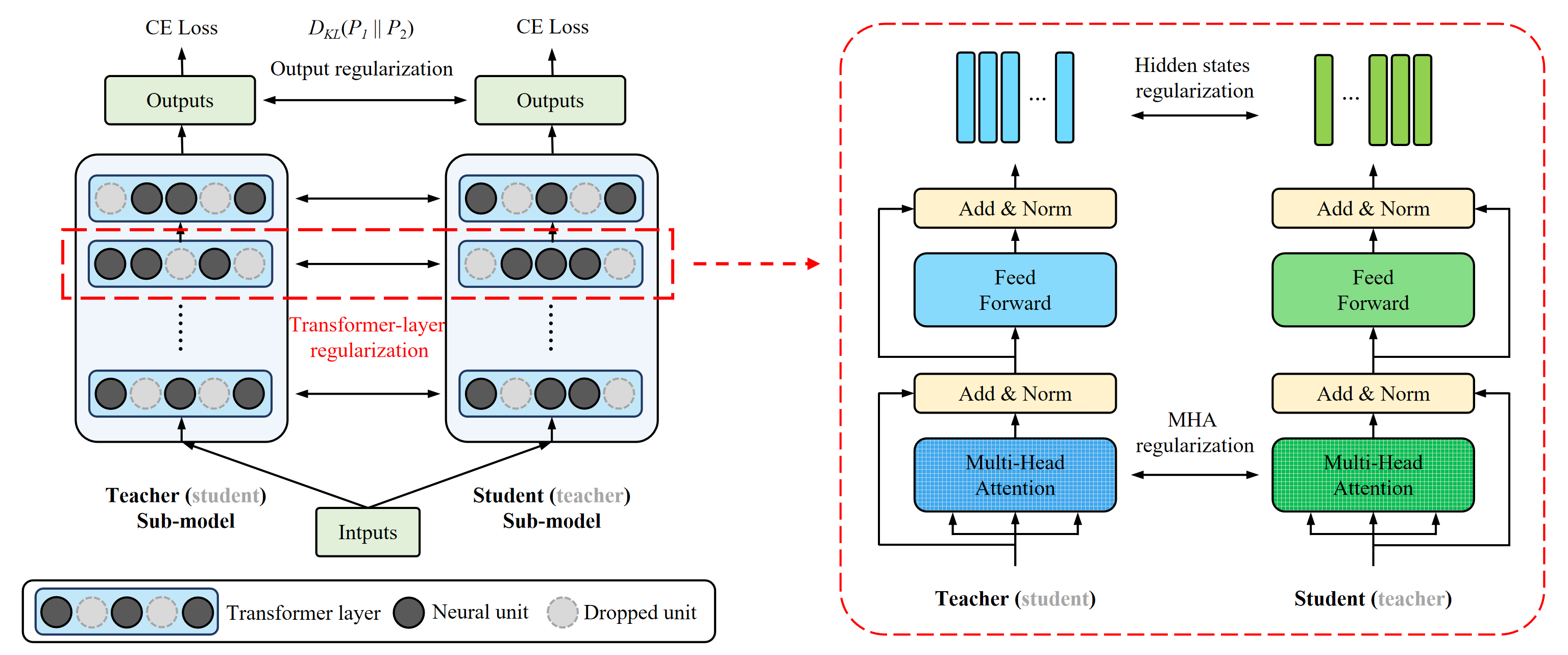}
	\caption{The proposed LR-Drop to regularize Transformer-based PLM. The left figure shows that one input will go through the two different sub-models produced by dropout twice and obtain two distributions $P_1$ and $P_2$. The right one shows a Transformer-layer regularization containing hidden states regularization MHA regularization.
	}
	\label{f1}
\end{figure*}
As shown in Figure 1, the red box is a Transformer-layer regularization, and we regularize between the two sub-Transformer-layers sampled in each layer of the model. The two sub-models obtained by dropout random sampling are mutually teacher and student. Therefore, Transformer-layer regularization can also be seen as Transformer-layer self-distillation. The right side of Figure 1 shows a concrete representation of the Transformer-layer regularization, which contains hidden states regularization and multi-head attention regularization. 

\textbf{Hidden States Regularization}. A fully connected feed-forward network is included in each transformer layer, which is expressed as follows:
\begin{equation}\label{key}
	\textbf{\textit{HS}}(x)= \mathrm{max}(0,x\textbf{W}_1+b_1)\textbf{W}_2+b_2,
\end{equation}
where there are two linear transformations and one ReLU activation in each feed-forward network. We regularize the knowledge from the Transformer layer outputs of the two sub-models with the following objectives: 
\begin{equation}\label{key}
	\mathcal{L}_{HSR}=\mathrm{MSE}(\textbf{\textit{HS}}(x)_1,\textbf{\textit{HS}}(x)_2),
\end{equation}
where the matrices $\textbf{\textit{HS}}(x)_1\in\mathbbm{R}^{l\times d}$ and   $\textbf{\textit{HS}}(x)_2\in\mathbbm{R}^{l\times d}$ are the hidden states of first sub-model and the second sub-model respectively, which are calculated by Equation 2. MSE() is the \textit{mean squared error} loss function, $l$ is the input text length, and $d$ is the hidden sizes of this two sub-models. 

\textbf{Multi-head Attention Regularization}. The key point for the Transformer-based model to work well is that each Transformer contains a multi-headed attention module whose attention function is computed depending on the query, key and value, represented as matrices $\textbf{\textit{Q}}$, $\textbf{\textit{K}}$ and $\textbf{\textit{V}}$. The attention function is called “Scaled Dot-Product Attention”, it can be expressed as:
\begin{equation}\label{key}
	\mathrm{Attention}(\textbf{\textit{Q}},\textbf{\textit{K}},\textbf{\textit{V}})=\mathrm{softmax}(\dfrac{\textbf{\textit{QK}}^T}{\sqrt{d_k}})\textbf{\textit{V}},
\end{equation}
where $d_k$ is the dimension of queries and keys. Then the dot product of the query and all the keys is obtained by dividing each key by $sqrt{d_k}$ and applying the softmax() function to obtain the weights of the values. Multi-head attention is concatenated by the attention of independent initialization weights in equation (4), which enables the model to focus jointly on information from different representation subspaces. It can be expressed as:
\begin{equation}\label{key}
	\mathrm{MHA}(\textbf{\textit{Q}},\textbf{\textit{K}},\textbf{\textit{V}})=\mathrm{Concat}(A^1_{head},...,A^2_{head})\textbf{\textit{W}}^o,
\end{equation}
where $h$ is the number of attention heads, and $A^1_{head}$ denotes the $i$-th attention head, which is calculated by the equation (4). The matrix $\textbf{\textit{W}}^o$ acts as a linear transformation.

Multi-head attention learns a lot of linguistic knowledge during training, and it is necessary to regularize it. Therefore, we propose multi-head attention regularization to encourage mutual learning of attention weights between two sub-models. The optimization objective is defined as:
\begin{equation}\label{key}
	\mathcal{L}_{MHAR}=\frac{1}{h}\sum^h_{i=1}\mathrm{MSE}(\textbf{\textit{A}}^1_i,\textbf{\textit{A}}^2_i),
\end{equation}
where $h$ is the number of attention heads, $\textbf{\textit{A}}^1_i\in\mathbbm{R}^{l\times l}$ and $\textbf{\textit{A}}^2_i\in\mathbbm{R}^{l\times l}$ refer to the attention matrix corresponding to the i-th head of first sub-model and the second sub-model, $l$ is the input text length, and MSE() refer to the \textit{mean squared error} loss function.
\subsection{Output Regularization}
In addition to regularizing the Transformer layer within the model, we also apply regularization to the output of the model, similar to R-Drop. Specifically, our approach, LR-Drop, aims to minimize the bidirectional KL-divergence between the output distributions of the two sub-models obtained through dropout sampling. The optimization objective is defined as:
\begin{equation}\label{key}
	\mathcal{L}_{OR}=\frac{1}{2}[\mathrm{KL}(P_1,P_2)+\mathrm{KL}(P_2,P_1)],
\end{equation}
where $P_1$ and $P_2$ are the output distributions of the first and second sub-models, respectively, and KL() denotes the KL-divergence loss function.

\subsection{Total Optimization Objective}

To summarize, the total optimization objective of our proposed LR-Drop during training is expressed as follows:
\begin{equation}\label{key}
	\mathcal{L}_{Total}=\mathcal{L}_{CE}+\alpha\mathcal{L}_{HSR}+\beta\mathcal{L}_{MHAR}+\gamma\mathcal{L}_{OR},
\end{equation}
where $\alpha$, $\beta$, and $\gamma$ are the weight coefficients for the regularization loss functions $\mathcal{L}_{HSR}$, $\mathcal{L}_{MHAR}$, and $\mathcal{L}_{OR}$, respectively.

\section{Experiments}
We assessed the effectiveness of LR-Drop on various natural language processing tasks. Our evaluation involved eight datasets for natural language understanding, six datasets for neural machine translation, and one dataset for abstractive summarization. In the table below, we use the abbreviations "RD" to refer to R-Drop and "LRD" to refer to LR-Drop in the presentation of the experimental results.
\begin{table*}[t]
	\centering
	\setlength\tabcolsep{6pt} 
	\begin{tabular}{lccccccccc}
		\toprule[1pt]
		\textbf{Model}               & \textbf{MNLI} & \textbf{MRPC} & \textbf{QNLI} & \textbf{QQP}  & \textbf{RTE}  & \textbf{SST-2} & \textbf{STS-B} & \textbf{CoLA} & \textbf{Avg}   \\\hline
		BERT-base                    & 83.8          & 85.3          & 90.8          & 91.0          & 68.2          & 92.4           & 89.3           & 62.3          & 82.85          \\
		BERT-base + RD               & 85.5          & 87.3          & 92.0          & 91.4          & 71.1          & 93.0           & 89.6           & 62.6          & 84.06          \\
		\textbf{BERT-base + LRD}     & \textbf{86.1} & \textbf{88.0} & \textbf{92.3} & \textbf{91.4} & \textbf{71.8} & \textbf{93.2}  & \textbf{90.2}  & \textbf{63.4} & \textbf{84.55} \\ \hline
		RoBERTa-large                & 90.2          & 90.9          & 94.7          & 92.2          & 86.6          & 96.4           & 92.4           & 68.0          & 88.93          \\
		RoBERTa-large + RD           & 90.9          & 91.4          & 95.2          & 92.5          & 88.4          & 96.9           & 92.5           & 70.0          & 89.73          \\
		\textbf{RoBERTa-large + LRD} & \textbf{91.3} & \textbf{91.8} & \textbf{95.9} & \textbf{93.2} & \textbf{89.2} & \textbf{97.4}  & \textbf{92.7}  & \textbf{71.8} & \textbf{90.41} \\ \hline
		ELECTRA-large                & 90.9          & 90.8          & 95.0          & 92.4          & 88.0          & 96.9           & 92.6           & 69.1          & 89.46          \\
		ELECTRA-large + RD           & 91.2          & 91.3          & 95.6          & 92.6          & 88.9          & 97.4           & 92.8           & 70.5          & 90.03         \\
		\textbf{ELECTRA-large + LRD} & \textbf{91.7} & \textbf{92.1} & \textbf{96.2} & \textbf{93.3} & \textbf{89.5} & \textbf{97.6}  & \textbf{93.1}  & \textbf{71.9} & \textbf{90.68} \\ \bottomrule[1pt]
	\end{tabular}
	\caption{Performances on natural language understanding tasks of GLUE benchmark. Significance test: the average performance of LR-Drop and R-Drop on the GLUE datasets was t-tested to obtain a p-value of 0.0034 < 0.01.}
\end{table*}
\subsection{Natural Language Understanding}
\textbf{Datasets}
We begin by assessing the effectiveness of LR-Drop on natural language understanding tasks. The GLUE Benchmark consists of eight English natural language understanding tasks, which vary in domains, data volumes, and difficulty levels. 

(1) \textbf{RTE} \cite{dagan2006pascal,bar2006second,giampiccolo2007third}: This dataset comprises a series of natural language inference datasets used in annual text challenges.

(2) \textbf{MNLI} \cite{williams2018broad}: In this task, a premise and a hypothesis are given, and the objective is to predict whether the premise supports or contradicts the hypothesis, or neither.

(3) \textbf{MRPC} \cite{dolan2005automatically}: Given a pair of sentences, the task is to determine whether their semantics are the same.

(4) \textbf{STS-B} \cite{agirre2007semantic}: Each data instance consists of a pair of sentences along with a similarity score ranging from 1 to 5. The task involves discrete regression to predict the scores.

(5) \textbf{QQP} \cite{qqp2016url}: This task involves identifying whether a pair of questions are semantically identical.

(6) \textbf{SST-2} \cite{socher2013recursive}: This binary task requires predicting whether a sentence is positive or negative.

(7) \textbf{QNLI} \cite{rajpurkar2016squad}: The objective of this task is to determine if a given question can be answered using the context sentence.

(8) \textbf{CoLA} \cite{warstadt2018neural}: This task focuses on assessing the grammatical accuracy of a sentence.

\textbf{Experimental Settings}
In this subsection, we employ three publicly available pre-trained language models (PLMs) as the baseline models for our experiments to evaluate the effectiveness of LR-Drop. The chosen PLMs are BERT-base, RoBERTa-large, and ELECTRA-large. 
Different tasks may require different parameter settings, so we dynamically adjust the coefficients $\alpha$, $\beta$, and $\gamma$ from the set $\{0.01, 0.05, 0.1, 0.5\}$ accordingly. The experimental methodology for the comparative models follows the approach outlined in previous research. 
For the STS-B task, we use the Pearson correlation as the evaluation metric, while for CoLA, we employ Matthew's correlation. The remaining tasks are evaluated based on Accuracy. We report the mean results of 5 runs to ensure statistical reliability. 
The experiments were conducted using an RTX 3090 GPU.

\textbf{Experimental Results}
The experimental results are presented in Table 1. When applying LR-Drop to the BERT-base model, we observed improvements in fine-tuning scores across multiple tasks. For the MNLI task, the fine-tuning score increased from 83.8 to 86.1. In the MRPC task, the score improved from 85.3 to 88.8. The QNLI task saw an improvement from 90.8 to 92.3, and the QQP task improved from 91.0 to 91.4. For the RTE task, the score increased from 68.2 to 71.8. The SST-2 task showed an improvement from 92.4 to 93.2, and the STS-B task improved from 89.3 to 90.2. Similarly, the RoBERTa-large and ELECTRA-large models also exhibited performance improvements of more than 1 point per dataset when using LR-Drop.

Across the eight datasets, the BERT-base + LR-Drop, RoBERTa-large + LR-Drop, and ELECTRA-large + LR-Drop models achieved impressive average scores of 84.55, 90.41, and 90.68, respectively. LR-Drop significantly improved the performance of the three baseline models: BERT-base, RoBERTa-large, and ELECTRA-large, by 1.70 points, 1.48 points, and 1.22 points, respectively. Furthermore, when compared to the previous method R-Drop, our proposed LR-Drop demonstrated average improvements of 0.49 points, 0.65 points, and 0.68 points, respectively. These results indicate that the performance of our LR-Drop regularization is particularly enhanced when applied to stronger baseline models. Additionally, the effectiveness of LR-Drop is evident across different neural language models, resulting in improved performance in natural language understanding tasks.

\begin{table*}[]
	\centering
	\setlength\tabcolsep{1pt}
	\begin{tabular}{lccccccccc}
		\toprule[1pt]
		\textbf{Model}               & \textbf{En to De} & \textbf{De to En} & \textbf{En to Fr} & \textbf{Fr to En} & \textbf{En to Zh} & \textbf{Zh to En} & \textbf{En to Es} & \textbf{Es to En} & \textbf{Avg}   \\ \hline
		Transformer                  & 28.57             & 34.64             & 35.9              & 36.1              & 26.3              & 18.4              & 39.0              & 40.6              & 32.44          \\
		Transformer + RD             & 30.72             & 37.25             & 38.0              & 38.9              & 28.1              & 19.5              & 41.8              & 43.2              & 34.68          \\
		\textbf{Transformer + LRD}   & \textbf{30.95}    & \textbf{37.87}    & \textbf{38.8}     & \textbf{39.6}     & \textbf{28.6}     & \textbf{20.3}     & \textbf{42.6}     & \textbf{44.1}     & \textbf{35.35} \\ 	\bottomrule[1pt]
	\end{tabular}
	\caption{BLEU scores on 8 IWSLT machine translation tasks. Significance test: the average performance of LR-Drop and R-Drop on the 8 datasets was t-tested to obtain a p-value of 0.0037 < 0.01.}
\end{table*}
\subsection{Neural Machine Translation}
\textbf{Datasets}
The datasets used for neural machine translation were obtained from the International Workshop on Spoken Language Translation (IWSLT) competitions. These datasets consist of translations between English and German (En $\leftrightarrow$ De), English and Spanish (En $\leftrightarrow$ Es), English and French (En $\leftrightarrow$ Fr), and English and Chinese (En $\leftrightarrow$ Zh). Specifically, we used the IWSLT14 dataset for English to German and vice versa, the IWSLT14 dataset for English to Spanish and vice versa, the IWSLT17 dataset for English to French and vice versa, and the IWSLT dataset for English to Chinese and vice versa. The IWSLT dataset contains approximately 170,000 pairs of sentences for training, 7,000 pairs for validation, and 7,000 pairs for testing. These datasets serve as valuable resources for training and evaluating our neural machine translation models.

\textbf{Experimental Settings}
The experimental configuration outlined in \citep{wu2021r} is followed in this study. Our benchmark model is the Transformer network proposed by \citep{vaswani2017attention}. The specific configurations for the IWSLT translations are specified under the transformer\_iwslt\_de\_en setting. To explore different settings, the coefficients $\alpha$, $\beta$, and $\gamma$ are dynamically varied within the set $\{0.1, 0.5, 1\}$. The implementation of our models is carried out using the Fairseq framework. For evaluating the performance of the models on neural machine translation tasks, we utilize BLEU scores. The reported results are the averages obtained from five trial runs to ensure robustness. The experiments are conducted on an RTX 3090 GPU, which serves as the hardware for the experiments.

\textbf{Experimental Results} The experimental results are presented in Table 2. When applying the Transformer model to our LR-Drop technique, we observed improvements in the fine-tuning scores for various translation tasks. For the English to German translation task, the fine-tuning score increased from 28.57 points to 30.95 points. Similarly, for the German to English translation, the score improved from 34.64 points to 37.87 points. For the English to French translation, the score improved from 35.9 points to 38.8 points, and for the French to English translation, the score improved from 36.1 points to 39.6 points. In the English to Chinese translation task, the score improved from 26.3 points to 28.6 points, and in the Chinese to English translation, the score improved from 18.4 points to 20.3 points. Additionally, for the English to Spanish translation, the score improved from 39.0 points to 42.6 points, and for the Spanish to English translation, the score improved from 40.6 points to 44.1 points.

Comparing the Transformer model with the LR-Drop technique to the original Transformer, we observed an average improvement of 2.91 points across the eight IWSLT machine translation tasks. Furthermore, the Transformer model with R-Drop achieved an average score of 34.68, while our Transformer model with LR-Drop achieved an average score of 35.35. These results demonstrate the effectiveness of our method in improving the performance of neural machine translation tasks.
\begin{table}[]
	\centering
	\setlength\tabcolsep{9.8pt}
	\begin{tabular}{llll}
		\toprule[1pt]
		\textbf{Model}      & \textbf{RG-1}  & \textbf{RG-2}  & \textbf{RG-L}  \\
		\hline
		Transformer$^a$         & 39.50          & 16.06          & 36.63          \\
		ProphetNet$^b$          & 44.02          & 21.17          & 41.30          \\
		BART$^c$                & 44.16          & 21.28          & 40.90          \\
		PEGASUS$^d$             & 44.17          & 21.47          & 41.11          \\
		BART + R3F$^e$          & 44.38          & 21.53          & 41.17          \\
		BART + RD$^f$           & 44.51          & 21.58          & 41.24          \\
		\textbf{BART + LRD} & \textbf{44.58} & \textbf{21.63} & \textbf{41.30} \\ 	\bottomrule[1pt]
	\end{tabular}
	\caption{The ROUGE scores, consisting of ROUGE-1, ROUGE-2, and ROUGE-L, are presented for the CNN/Daily Mail summarization dataset. A significance test was carried out, comparing the mean performance of LR-Drop and R-Drop. A t-test yielded a p-value of 0.0064, which is less than 0.01. References: $^a$:~\citep{vaswani2017attention}; $^b$:~\citep{qi2020prophetnet}; $^c$:~\citep{lewis2020bart}; $^d$:~\citep{zhang2020pegasus}; $^d$:~\citep{aghajanyan2020better}; $^f$:~\citep{wu2021r}.}
	\label{tab:my-table}
	\vspace{-0.5em}
\end{table}
\begin{table*}[t]
	\centering
	\setlength\tabcolsep{6.8pt} 
	\begin{tabular}{lccccccccc}
		\toprule[1pt]
		\textbf{Model}               & \textbf{MNLI} & \textbf{MRPC} & \textbf{QNLI} & \textbf{QQP}  & \textbf{RTE}  & \textbf{SST-2} & \textbf{STS-B} & \textbf{CoLA} & \textbf{Avg}   \\\hline
		LR-Drop     & 86.1 & {88.0} & {92.3} & {91.4} & {71.8} & {93.2}  & {90.2}  & {63.4} & {84.55} \\ \hline
		~~-- w/o $\mathcal{L}_{HSR}$             & 85.7&	87.4	&92.2&	91.4&	71.6&	93.1&	89.6&	62.7&	84.21 
		\\
		~~-- w/o $\mathcal{L}_{MHAR}$           & 85.9          & 87.4          & 92.2          & 91.6          & 71.7          & 93.1           & 89.9           & 62.7          & 84.30          \\
		~~-- w/o $\mathcal{L}_{OR}$           & 85.8          & 87.5          & 92.1          & 91.5          & 71.5          & 93.2           & 89.7           & 62.8          & 84.26          \\
		\bottomrule[1pt]
	\end{tabular}
	\caption{Ablation study of LR-Drop.}
\end{table*}
\begin{table*}[]
	\centering
	\setlength\tabcolsep{14pt}
	\begin{tabular}{l|c|c|c|c|c|c}
		\toprule[1pt]
		\multirow{2}{*}{Methods} & \multicolumn{6}{c}{size of the training set} \\ \cline{2-7} &1K & 2K & 4K & 8K & 16K & 32K \\ \hline 
		BERT-base &53.10 & 54.24 & 57.34 & 62.39 & 86.58 & 89.16 \\
		BERT-base + R-Drop &54.12 & 56.21 & 57.87 & 62.74 & 87.10 & 89.56 \\
		BERT-base + LR-Drop &55.04 & 56.53 & 58.56 & 63.47 & 88.03 & 90.07\\  \bottomrule[1pt]
	\end{tabular}
	\caption{\small LR-Drop performance on training sets of different sizes. Dataset: SST-2, baseline model: BERT-base.}
	\label{t5}
\end{table*}
\subsection{Abstractive Summarization}
\textbf{Datasets}
The abstract summarization task aims to condense lengthy sentences or documents into concise sequences while preserving the main content. It is a generation task. To evaluate the effectiveness of LR-Drop on the abstract summarization task, we employ the CNN/Daily Mail dataset. This dataset consists of news documents (source) and their corresponding highlighted summaries (target), which are extracted from CNN and Daily Mail websites. The dataset comprises 287,226 training documents, 13,368 validation documents, and 11,490 test documents. We preprocess the dataset following the guidelines provided by Wu et al. (2021) \citep{wu2021r}.

\textbf{Experimental Settings}
In this subsection, we conduct experiments using the BART pre-training model as a baseline to assess the efficacy of our proposed LR-Drop regularization technique. During the fine-tuning process, we apply the BART model with LR-Drop. The coefficients $\alpha$, $\beta$, and $\gamma$ are set to 0.1, 0.2, and 0.5, respectively. The remaining hyper-parameter settings are consistent with the original paper on the BART model. To evaluate the performance of the models on the abstract summarization task, we employ the ROUGE F1 score. The reported results are the mean scores obtained from five independent runs. The experiments are conducted on an RTX 3090 GPU.

\textbf{Experimental Results}
To provide a comprehensive evaluation of the different methods, we report the ROUGE-1 (RG-1) and two-sequence ROUGE-2 (RG-2) overlaps to assess the information content, as well as the longest common subsequence ROUGE-L (RG-L) scores to evaluate fluency. The experimental results are presented in Table 3. Notably, BART, when trained with our LR-Drop technique, outperforms all other models and achieves state-of-the-art performance. Compared to the original BART model, BART + LR-Drop demonstrates improvements of 0.41 points in RG-1, 0.35 points in RG-2, and 0.40 points in RG-L scores. Additionally, BERT + LR-Drop surpasses BERT + R3F and BERT + R-Drop. These experimental findings highlight the effectiveness of our LR-Drop technique in enhancing the performance of abstract summarization tasks. 

\section{Study and Analysis on LR-Drop}
In this section, we present a comprehensive investigation and analysis of LR-Drop, considering multiple aspects such as ablation study, training set sizes, dropout times, and loss landscape analysis. All experiments in this section are conducted using the datasets from the General Language Understanding Evaluation (GLUE) benchmark.
\subsection{Ablation Study}
To further analyze the role of each component in LR-Drop, we conducted ablation experiments by removing individual parts of the loss function: $\mathcal{L}_{HSR}$, $\mathcal{L}_{MHAR}$, and $\mathcal{L}_{OR}$. This allowed us to assess the contribution of each component to the overall performance of LR-Drop. The following three ablated methods were obtained:
(1) LR-Drop w/o $\mathcal{L}_{HSR}$: This method removes the hidden states regularization component from LR-Drop.
(2) LR-Drop w/o $\mathcal{L}_{MHAR}$: This method removes the multi-head attention regularization component from LR-Drop.
(3) LR-Drop w/o $\mathcal{L}_{OR}$: This method removes the output regularization component from LR-Drop.

Table 4 presents the results of the ablation experiments. It is evident that the full LR-Drop method achieves the best performance. When the hidden states regularization component is ablated, the average score of LR-Drop on GLUE decreases to 84.21, which is 0.34 points lower than the full LR-Drop. Similarly, when the multi-head attention regularization component is removed, the average score of LR-Drop on GLUE decreases to 84.30, which is 0.25 points lower than the full LR-Drop. Lastly, when the output regularization component is ablated, the average score of LR-Drop on GLUE decreases to 84.26, which is 0.29 points lower than the full LR-Drop.

These results highlight the importance of each component in LR-Drop and demonstrate that the combination of all components leads to the highest performance on GLUE tasks.

\subsection{LR-Drop Performance on Training Sets of Different Sizes}
This study also investigates the performance of LR-Drop across training datasets of varying sizes. To examine this, we partitioned the SST-2 training set into different subsets and utilized the BERT-base model with the same structure as previously described. The results of this experiment are presented in Table \ref{t5}, where we compare LR-Drop with R-Drop.
As shown in Table \ref{t5}, our approach, LR-Drop, not only achieves superior performance on large training datasets but also demonstrates significant improvements on smaller datasets. In comparison to the BERT-base model, our enhanced model, BERT-base + LR-Drop, achieves performance boosts of 1\% to 2\% across distinct training sets. Furthermore, LR-Drop outperforms R-Drop on training sets of various sizes. These experimental findings highlight the effectiveness of LR-Drop in providing robust regularization for the model, even when data availability is limited.

\begin{table*}[t]
	\centering
	\setlength\tabcolsep{9pt} 
	\begin{tabular}{lccccccccc}
		\hline
		\textbf{Model}               & \textbf{MNLI} & \textbf{MRPC} & \textbf{QNLI} & \textbf{QQP}  & \textbf{RTE}  & \textbf{SST-2} & \textbf{STS-B} & \textbf{CoLA} & \textbf{Avg}   \\\hline
		LR-Drop-2     & 86.1 & {88.0} & {92.3} & {91.4} & {71.8} & {93.2}  & {90.2}  & {63.4} & {84.55} \\ \hline
		LR-Drop-3   & 86.2 & 88.1 & 92.3 & 91.3 & 71.8 & 93.4  & 90.1  & 63.4 & 84.58    \\
		\hline
	\end{tabular}
	\caption{The effect of the number of LR-Drop samples on performance.}
\end{table*}
\begin{figure*}[]
	\centering
	\includegraphics[width=0.999\linewidth]{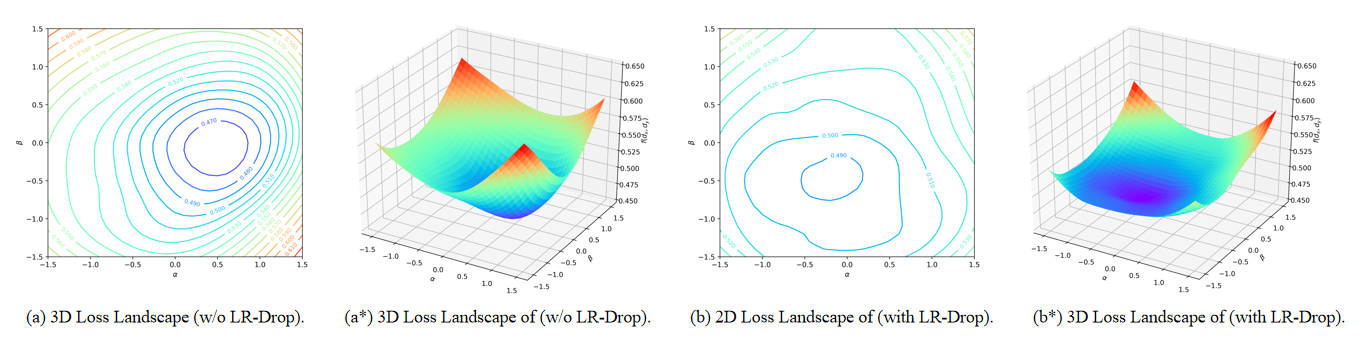}
	\caption{2D (left) and 3D (right) visualization of loss function minima selected by BERT-base with standard training (ST) and LR-Drop on SST-2 dataset. 
	}
	\label{f5}
\end{figure*}
\subsection{3-time LR-Drop}
From the above, it is evident that LR-Drop is based on the concept of "twice dropout" to generate two sub-networks for regularization during training. In our research, we extended this approach by increasing the number of forward propagations from two to three, meaning that each example passes through the model three times in LR-Drop. During training, we also incorporate hidden states regularization, multi-head attention regularization, and output regularization to facilitate mutual learning and knowledge exchange among the three sub-models.
The experimental results, as presented in Table 6, demonstrate that LR-Drop-3, which consists of three sub-models, yields a slight improvement of 0.03 points over LR-Drop-2, the original LR-Drop with two sub-models. However, it is worth noting that increasing the number of dropout samples tends to increase computational consumption. Therefore, we believe that utilizing LR-Drop-2 regularization twice is sufficient for achieving effective model training. However, if computational resources are not a constraint, one can consider using LR-Drop-3.

It is important to mention that an increase in iterations corresponds to an increase in resource usage. Therefore, we are of the view that the double iteration approach in LR-Drop-2 provides adequate regularization during model training. However, if computational resources are not a limitation, one might explore the option of deploying LR-Drop-3.
\subsection{Loss Landscape Analysis}
To provide a more visual analysis of the regularization effect of LR-Drop, we employed a method proposed by Li et al. (2018) to visualize high-dimensional non-convex loss functions. This method allowed us to graphically depict the loss landscapes surrounding the empirical risk minima obtained through standard training and LR-Drop training, both with the same model structure. Figure \ref{f5} presents 2D and 3D visualizations of these loss landscapes.

In this visualization, we considered the dimensions of $\theta$ and randomly extracted two direction-representing vectors, $d_x$ and $d_y$, from a Gaussian distribution with a mean of zero. The scale of these vectors was set to be equivalent to the variance of the weight values. We then introduced linear perturbations $\alpha$ and $\beta$ to a central point $\theta^\ast$, resulting in a loss function that represents variations in these two random directions:
\begin{equation}\label{key}
	f(d_x,d_y)=\mathcal{L}(\theta^\ast+\alpha d_x+\beta d_y)
\end{equation}

The analysis points out that the proposed LR-Drop indeed selects flatter loss landscapes by dynamically creating perturbation. Numerous investigations have evidenced that a flatter loss landscape generally implies superior generalization \citep{keskar2019large,ishida2020we}.

\section{Conclusion}
In this work, we have introduced LR-Drop, a simple yet effective regularization technique based on dropout. LR-Drop utilizes layer-wise self-distillation between two sub-models generated by dropout to improve model performance. During training, LR-Drop regularizes different representations in Transformer-based Pre-trained Language Models (PLMs), including the hidden state, the multi-headed attention matrix, and the output distribution of the prediction layer. Importantly, LR-Drop does not introduce additional model parameters or modify the original architecture of the language model, making it applicable to various Transformer-based PLMs.
Through extensive experiments on a diverse set of 15 datasets, including natural language understanding, neural machine translation, and abstractive summarization tasks, we have demonstrated that LR-Drop achieves superior performance compared to existing methods. Our results even surpass state-of-the-art performance in these tasks. This highlights the effectiveness and versatility of LR-Drop as a regularization technique for Transformer-based PLMs.  
\section{Acknowledgements}
This research was funded by the State Sponsored Postdoctoral Researcher Program of China. Grant No. GZC20232873.
\section{Bibliographical References}\label{sec:reference}

\bibliographystyle{lrec-coling2024-natbib}
\bibliography{lrec-coling2024-example}

\end{document}